\documentclass[preprint,12pt,authoryear]{elsarticle}

\usepackage{amssymb}
\usepackage{gensymb}
\usepackage{graphicx}
\usepackage{xcolor}
\usepackage{tikz}
\usepackage{hyperref}
\usepackage{multirow}  
\usepackage{booktabs}       

\usetikzlibrary{arrows}
\usetikzlibrary{shapes}
\usetikzlibrary{backgrounds}
\begin{document}

\begin{frontmatter}



\title{Mixed Evidence for Gestalt Grouping in Deep Neural Networks}


\author[inst1]{Valerio Biscione\corref{cor1}}

\cortext[cor1]{Corresponding Author: valerio.biscione@bristol.ac.uk}

\affiliation[inst1]{organization={Department of Psychology},
            addressline={University of Bristol}, 
            city={Bristol},
            postcode={BS8 1TL}, 
            country={United Kingdom}}

\author[inst1]{Jeffrey S. Bowers}



\begin{abstract}

Gestalt psychologists have identified a range of conditions in which humans organize elements of a scene into a group or whole, and perceptual grouping principles play an essential role in scene perception and object identification. Recently, Deep Neural Networks (DNNs) trained on natural images (ImageNet) have been proposed as compelling models of human vision based on reports that they perform well on various brain and behavioral benchmarks.  Here we test a total of 16 networks covering a  variety of architectures and learning paradigms (convolutional, attention-based, supervised and self-supervised, feed-forward and recurrent) on dots (Experiment 1) and more complex shapes (Experiment 2) stimuli that produce strong Gestalts effects in humans. In Experiment 1 we found that convolutional networks were indeed sensitive in a human-like fashion to the principles of proximity, linearity, and orientation, but only at the output layer. In Experiment 2, we found that most networks exhibited Gestalt effects only for a few sets, and again only at the latest stage of processing. Overall, self-supervised and Vision-Transformer appeared to perform worse than convolutional networks in terms of human similarity. Remarkably, no model presented a grouping effect at the early or intermediate stages of processing. This is at odds with the widespread assumption that Gestalts occur prior to object recognition, and indeed, serve to organize the visual scene for the sake of object recognition. Our overall conclusion is that, albeit noteworthy that networks trained on simple 2D images support a form of Gestalt grouping for some stimuli at the output layer, this ability does not seem to transfer to more complex features. Additionally, the fact that this grouping only occurs at the last layer suggests that networks learn fundamentally different perceptual properties than humans.


\end{abstract}




\end{frontmatter}
\tnotetext[1]{}

\section{Introduction} \label{Introduction}

Human tends to group perceptual features together in order to form a coherent whole. Understanding when this happens has been the focus of Gestalt psychology research for over 100 years and more than a hundred grouping ``laws'' have been suggested \citep{aCenturyI}. Whereas in the past the formulation of these laws was based on subjective experience and was criticised for a lack of scientific rigour, subsequent researchers have developed experimental designs with carefully constructed stimuli (e.g. Gabor stimuli, dot lattices) that allow for parametric control, richer visual displays, and objective measures of grouping effects \citep{aCenturyII}. One such approach consists of measuring the impact of salient Emergent Features (EFs) on discriminating visual patterns. These EFs derive from the relationship amongst individual parts rather than the parts themselves \citep{Pomerantz1977PerceptionEffects, PomerantzPortillo}. We will use the concept of EFs as the basis of our approach, as detailed later.

Recently there has been an explosion of interest in Deep Neural Networks (DNNs) as models of the human visual system for object recognition. Even though DNNs have primarily been designed to solve engineering tasks, reports that the pattern of activations of units in DNNs are similar to neural activation in human and macaque visual systems have led to the view that DNNs can be used as a test bed for simulating biological vision in mammals \citep{Gauthier2016, Kriegeskorte}. As a way of formalizing this similarity, a Brain-Score benchmark has been put forward \citep{SchrimpfBrainScore2018}, which has been enthusiastically embraced by researchers comparing DNNs to human vision.

In the current paper, we explore several DNNs thought to be amongst the best model of human vision and test whether they support various Gestalt grouping phenomena. In particular, we tested whether DNNs are sensitive to some basic principles of organization such as proximity, orientation, and linearity (Experiment 1), and whether they experience Gestalt grouping when presented with more complex stimuli (Experiment 2). We compare networks' responses with human responses from classic visual perception work \citep{Pomerantz1977PerceptionEffects, PomerantzPortillo}.

Our main research question is: do DNNs exhibit human-like Gestalt grouping effects? We split this question into two sub-questions: are networks sensitive to the basic properties of proximity, orientation and linearity?; are they sensitive to more complex Gestalt grouping effects? In addition to the primary aim of this investigation, we will also gain insight into what architecture or training regimes appear to be most appropriate for acquiring Gestalt properties, and to the extent to which Gestalt phenomena are a learned or innate aspect of the human visual system.





In the following sections, we contextualize these questions and explain how our experiments address them.

\subsection{Neural Networks as a Model of the Human Visual System} \label{IntroNNvisualsystem}
DNNs trained on ImageNet (a dataset consisting of 1000 categories of objects taken across over 1 million photographs, \citealt{ImageNet}) develop a set of internal feature representations that are statistically similar to the neural representations in human and non-human primate visual systems \citep{YaminsDiCarlo2016, Khaligh-Razavi2014, SchrimpfBrainScore2018}. A neuronal and behavioural benchmark called Brain-Score has been developed to assess neural networks on their similarity with biological object recognition systems, with DNNs performing much better than all previous approaches \citep{SchrimpfBrainScore2018}. At the time of writing more than 160 models have been tested on Brain-Score.

In spite of these successes when compared with neuronal data and tested on classification accuracy benchmarks, DNNs often fail on the most basic perceptual properties exhibited by humans \citep{BowersEtAl2022}. For example, DNNs do not possess human-like shape bias \citep{Geirhos2020, Malhotra2020}, they appear to discriminate categories based on local instead of global features \citep{Baker2018, Malhotra2022Plos}, 
are much more susceptible to a low amount of image degradation \citep{Geirhos2020}, do not account for humans' similarity judgments of 3D shapes \citep{German2020CanJudgments}, and fail to support basic visual reasoning such as classifying images as the same or different \citep{Puebla2021}. In addition, DNNs often act in surprising non-human-like ways, such as being fooled by adversarial images \citep{Adversarial2013, Dujmovic2020} and make bizarre classification errors to familiar objects in unusual poses \citep{Kauderer-Abrams2017a, Gong2014, Chen2017}. Furthermore, recent work by \cite{Xu2021} failed to find strong neural correlates between DNNs' internal representation and fMRI signals from high-level visual areas of human participants.


At the same time, some authors have highlighted that DNNs appear to capture some key psychological findings, such as Weber's Law, sensitivity to scene incongruencies, and the Thatcher effect \citep{Jacob2021} (but see \citealt{BowersEtAl2022} for limitation of these findings). \cite{Biscione2021JMLR, Biscione2022NN} also found that networks exhibited strong invariance to several novel object transformations (rotation, scale, change in luminance, translation, and to a lesser degree change in viewpoint), but only after being trained on a correspondingly transformed dataset of different classes, indicating that DNNs can learn the human perceptual property of object invariance to object transformation \citep{BlythingCommentary2021, Blything2021JoV}. Recently, \cite{DongYin2023} found that DNNs trained on words provided a better account for visual priming effects than many classic orthographic coding schemes.  

\subsubsection{Neural Networks and Gestalt} \label{IntroNNandGestalt}
As far as we are aware, Gestalt grouping in DNN have been explored only in relation to illusory contours and closure. \cite{Baker2018CogSci} tested the degree to which networks could  perceive  illusory  contours  after  being  trained  on  non-illusory  similar shapes  (large  and  thin  rectangles).  The  network successfully  predicted the type of shape regardless of whether the contour was normal or illusory, but the authors found no evidence that the network used the same information as humans and concluded that CNNs do not perceive illusory contours.  
\cite{Kim2019} challenged this conclusion and found that several architectures (including InceptionNet) pretrained on ImageNet exhibited closure on display of edge fragments. Whether the later findings reflect Gestalt processes similar to human visual processing remains unclear. Other researchers have focused on modifying architecture or training regime in order to explore this issue: \cite{Lotter2020} found that a network based on predictive coding (PredNet, one of the 16 networks tested in this work) which was trained on predicting the next frame of video sequences, exhibited disparate phenomena observed in the visual cortex, including the flash-lag effect and illusory contours. The same network also appeared to perceive illusory motion \citep{Watanabe2018}. Using a DNN modified to include feedback connections through predictive coding dynamics, \cite{Pang2021predcod} also showed human-like perception of illusory contours. 
Illusory contours constitute an important phenomenon in Gestalt psychology, but human perception is mediated by a wider set of grouping principles which, to the best of our knowledge, have not been explored in DNNs. 

If DNNs are going to be used as models of human vision it will be important that they not only do well on various Brain-Score measures but also account for key experimental results reported in psychology \citep{BowersEtAl2022}. Here we have focused on Gestalt rules of organization not only because they play a key role in visual perception and object recognition \citep{Biederman1987, Perrett1993, Spillman2009}, but because there are existing image datasets and robust empirical phenomena that make it easy to test.  Specifically, we consider whether DNNs show sensitivity to EFs, as described next.

\subsection{Formation of Wholes through Emergent Features} \label{IntroEFs}
Gestalt researchers have long studied the emergence of ``wholes" from the combination of individual parts, but Gestalts have proven difficult to define and measure. \cite{Pomerantz1977PerceptionEffects} operationally defined Gestalts as the result of salient Emergent Features (EFs), that is features that are the result of the relations amongst individual elements, and are not possessed by the elements themselves. These EFs behave as though they were elementary themselves and are sometimes detected more quickly than the more basic features from which they arise. The effect can be measured in a discrimination task, is robust to added noise \citep{CScontrast2020},  and might be a fundamental stage in shape processing related to the perception of non-accidental properties involved in shape perception as posed by the Recognition-by-Components theory\citep{Biederman2001RecognizingTheory, linesKubi2017}.

\begin{figure}
\centering
  \includegraphics[width=1\linewidth]{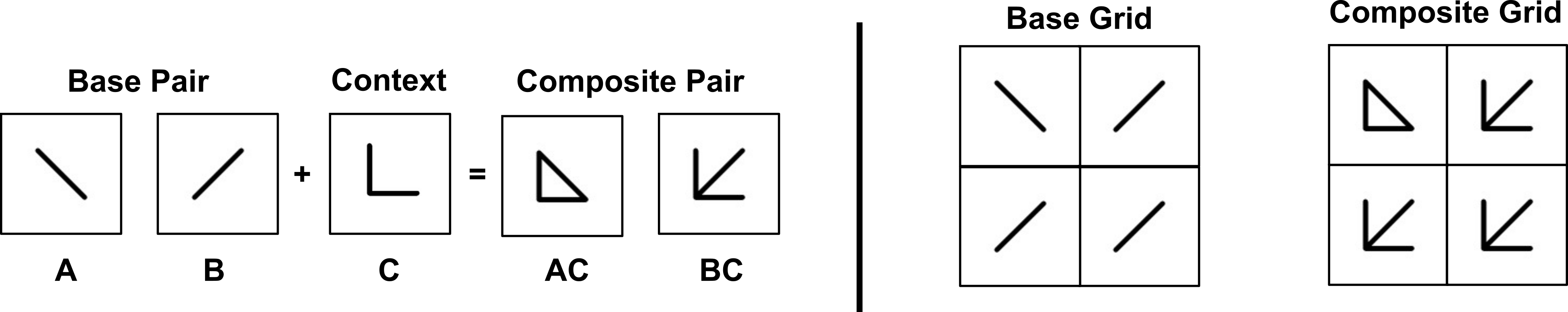}
\caption{\cite{Pomerantz1977PerceptionEffects} approach to measure Emergent Features (EFs). \textbf{Left}: starting with a base pair, a non-informative context is added to obtain a composite pair. \textbf{Right}: The base pair and composite pair are arranged in a 2x2 grid to form an odd-discrimination task. Participants are asked to indicate the location of the ``odd" element, and RTs are measured for the base and the composite grid. With some composite grids (such as the one illustrated in the figure) discrimination is much faster than the corresponding base grid. Since the added context in the composite is equal for odd and non-odd elements, any facilitation must be due to EFs.}
  \label{fig:1Pomsetup}
  \end{figure}

In Figure \ref{fig:1Pomsetup}, left, we illustrate this idea. As a baseline, we measure how well humans distinguish two stimuli (A and B, \emph{base pair}). We then add a new \emph{contextual} stimulus C to \emph{both} the A and B images, creating a \emph{composite} pair. Importantly, stimulus C is not informative by itself in distinguishing AC from BC. Nevertheless, the composite pair is now much more discriminable than the base pair due to the interaction of the context with the base elements (that is, through the creation of Emergent Features).
Whenever the context stimulus changes performance on the discrimination task a \emph{Configural Effect} (CE) is observed. Improved performance is described as a \emph{Configural Superiority Effects} (CSE) and this provides a measure of an EF that is the product of a Gestalt organizational principle.  By contrast, impaired performance is described as a \emph{Configural Inferiority Effect} (CIE), and does not reflect an EF, but rather, reflects a number of possible factors, including additional computational and attentional load, increase similarity, and crowding. That is, for a CSE to manifest itself, the EFs need to be powerful enough to override all these other factors. In a standard procedure, these stimulus sets have been used in a ``oddity reaction time (RT) task" where subjects were asked to determine in which quadrant of a 2x2 grid an ``odd" stimulus was presented (Figure \ref{fig:1Pomsetup}, right). Comparing RTs amongst each base and composite pair allows for the measurement of CSE and CIE in various contextual configurations. 

The advantage of this approach is that it provides a quantitative measure of CSEs/CIEs  rather than relying on subjective judgements. As predicted, many configurations (combination of characters, line segments forming letters, surfaces, 3D volumes) result in CIEs or very modest CSEs \citep{PomerantzPortillo}.  However, critically, other configurations show strong CSEs  \citep{Pomerantz1977PerceptionEffects, Pomerantz1989EmergentPerception, PomerantzPortillo}. In Experiment 1 we test network sensitivity to specific and low-level EFs (proximity, orientation, and linearity) with simple dot configuration following \cite{PomerantzPortillo}. The combination of these and others EFs (and possibly other factors,  including high-level features such as shape familiarity or closure) are assumed to give rise to the strong CSEs observed with the complex stimuli used in \cite{Pomerantz1977PerceptionEffects}. In Experiment 2, networks are tested on these stimuli as well as stimuli that produced strong CIEs in humans.

\subsection{Where do the laws of perception come from?} \label{IntroGestaltComeFrom}
A related question is the role of perceptual experience in acquiring EFs, that is, to what degree the grouping principles are learnt from the visual environment and to what degree they are a function of the innate architecture of the visual system \citep{Todorovic2011}. Classic Gestalt psychologists minimized the significance of a learning account \citep{Metzger1966}.  These authors conceded that some aspects of visual organization could be based on habit or learning (such as the ability to group particular continuous lines on a paper in letters), but these cases were thought to be the exception and weaker than others \citep{Wertheimer1923}.


Nevertheless, some evidence has emerged supporting the idea that basic Gestalt principles could be the result of learning from the visual environment: \cite{Geisler2001} showed that human performance in contour detection is quantitatively predicted by a grouping rule derived directly from the statistics of natural images; \cite{Peterson1994ObjectContours} found that a silhouette is more likely to be assigned as the figure if it suggests a common object (see also \citealt{Peterson2019}); with two different paradigms, \cite{Duncan1984} and \cite{Zemel2002} found that a perceptual grouping can be altered with only a small amount of experience in a novel stimulus environment. Other evidence based on RT responses has been collected by \cite{VeceraFarah1997}. However, some other combinations are impenetrable to learning, as demonstrated by the  Kanizsa stratification images \citep{Grossberg2017}.

In the current work, all networks tested were pre-trained on a large dataset of natural images: either ImageNet (a dataset consisting of a thousand categories of objects taken across over 1 million photographs, \citealt{ImageNet}) or, in the case of PredNet, the KITTI dataset (\citealt{KITTI}, a car-mounted camera dataset). Most of the networks used here achieve an impressive degree of accuracy on a test set of objects, at par with human performance on ImageNet \citep{He2015delvingDeep}. Therefore, regardless of their plausibility as a model of the human ventral pathways, we can use them to test whether learning statistical regularities on a complex domain of 2D natural images affords the network to extract EFs. Some of these regularities might be low-level such as being sensitive to the proximity of two stimuli, or their orientations with respect to one another; others might be higher level such as grouping based on shape familiarity. If we fail to observe grouping principles in our experiments it could be that the models need to be trained on more complex datasets (e.g. an interactive 3D world) in order for some Gestalt principles to emerge, or alternatively, different architectures are required.

\subsection {Outline of the current work} \label{IntroOutline}
We test a wide variety of DNNs (details in Section \ref{method networks}) on several sets of stimuli. Instead of using the ``odd" quadrant task, we presented each image to the neural network, and compared its internal activation across sets of images, always presented separately from one another. Each set is composed of two pairs of images, a base and a composite pair (obtained by adding a non-informative context to the base pair as in Figure \ref{fig:1Pomsetup}). For each pair, we computed the Euclidean distance between the activation vector at each networks' layer produced by the images of the pair. We then compared the Euclidean distances between the composite and the base pair and normalize it from -1 to 1 to obtain a Network Configural Effect (CE). Most of the results in the main text will refer to the Euclidean-based approach, but we also repeated the same analysis using the cosine similarity metric, a measure that is independent of the vectors' magnitude, which produced slightly different results as discussed in Section \ref{DiscussCosineSim}. The two metrics are outlined in detail in Section \ref{MethodMetrics} in \nameref{Methods}.

A positive CE is a measure of enhanced discrimination (Configural Superiority Effect, CSE); a negative of impaired discrimination (Configural Inferiority Effect, CIE). These measures can then be compared with CSEs/CIEs found in human participants assessed through reaction times (RTs) recording. While we compared humans and networks on both CIEs and CSEs, notice that only CSEs are the result of Gestalt grouping, while CIEs correspond to crowding/attention load, etc. We computed the CEs across the Convolutional and Fully Connected layers of the networks (before the non-linear operation was applied), with a particular emphasis on the output layer, as it appears to be the most appropriate for comparison with human RTs. Nevertheless, we expect CSEs to start emerging from earlier layers than the output layer given that Gestalt grouping is widely assumed to organize the visual scene for the sake of object recognition \citep{Biederman1987}.

We selected the models based on their historical importance, their performance on standard datasets, their biological plausibility and their Brain-Score (see the score for each network in \ref{networksScore}), and we aimed at using a wide variety of architectures and training regimes.  We used a total of 16 networks, all pre-trained on ImageNet (a part from PredNet which was pretrained on the KITTI dataset): 5 of them are classic convolutional networks, 3 are CNNs that have a direct biological inspiration  with a front-end that simulates primate V1, 4 are attention-based networks, 4 are self-supervised networks. Amongst all the above networks, 3 have recurrent mechanisms.  The human RTs data were collected from two sources: \cite{Pomerantz1977PerceptionEffects} and \cite{PomerantzPortillo}.
In Experiment 1, we investigated specific, low-level emergent features, by generating a wide number of configurations composed of simple dot patterns, as introduced by \cite{PomerantzPortillo} and illustrated in Figure \ref{fig:Exp1Stimuli}.  \cite{PomerantzPortillo} found a strong and consistent effect for three EFs: proximity, orientation, and linearity, and therefore we tested whether these same features could be used by the networks to enhance discriminability of a base image pair.

A multitude of low-level features combined together form Gestalt perceptual grouping for more complex shapes. In Experiment 2 we used 17 sets of stimuli that, albeit still simple, are composed of combinations of line segments and thus intend to elicit a wide set of emergent features in humans (\citealt{Pomerantz1977PerceptionEffects}, the whole set is shown in \ref{fig:Exp2corr}, left). Five of these sets generated high CSEs in humans, indicating a strong Gestalt grouping effect. It is assumed that the CSE observed in this complex stimuli is the result of many, low-level emergent features. In addition, another five sets generated high CIEs in humans, indicating a strong impact of a combination of attention load, crowding, etc. Both CSE and CIE sets are shown in the legends in Figures \ref{fig:Exp2_boxplot} and \ref{fig:Exp2_all_layers}. We generalized the results across different background conditions and transformation conditions (rotation, translation, scale, and no transformation).
More details about the networks used, and the analysis are presented in \nameref{Methods}. 

\section{Experiment 1} \label{Exp1}
\subsection{Methods}
\begin{figure}[!hb]
\centering
  \includegraphics[width=1\linewidth]{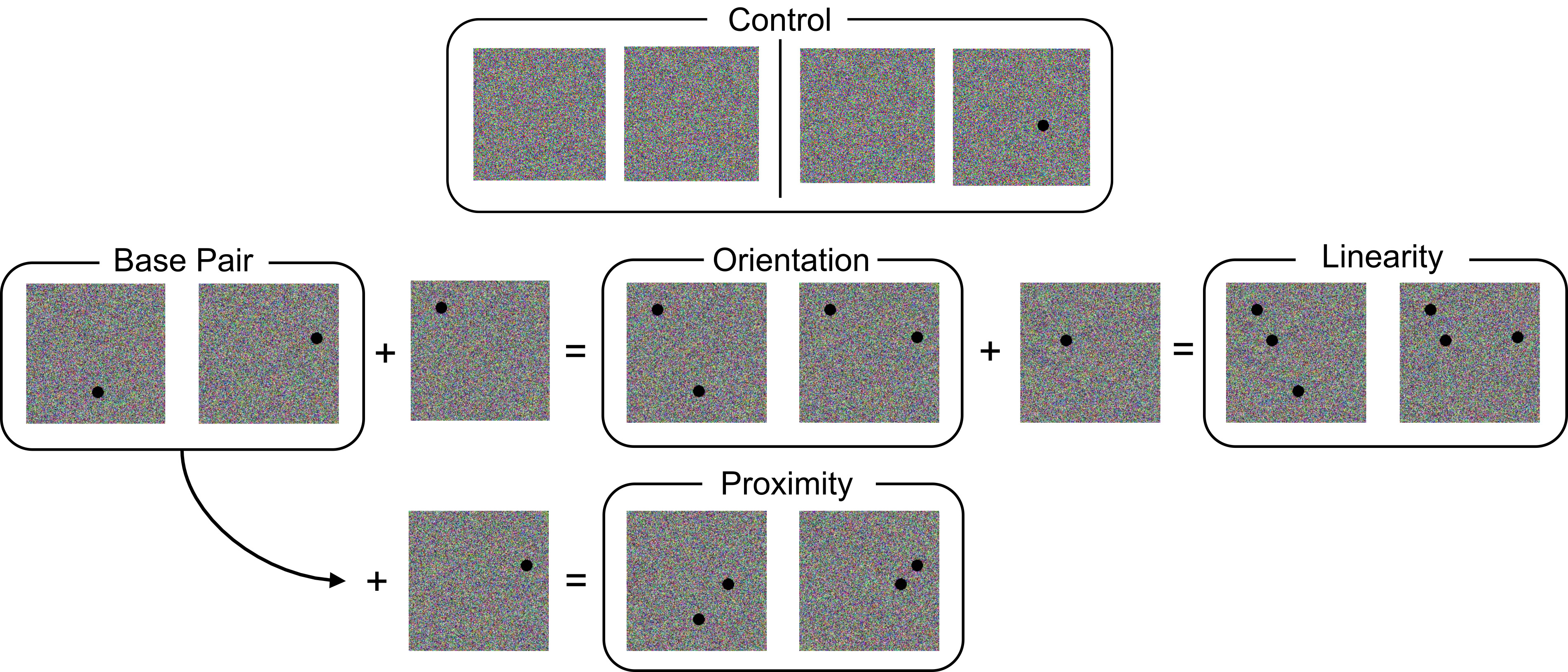}
\caption{Generation of stimuli for Experiment 1, following \cite{PomerantzPortillo}. \textbf{Top}: Control set: one pair always consists of randomly pixellated canvas without any stimuli on top; the other pair is obtained by adding one dot in a random location to one of the canvasses. \textbf{Bottom}: Starting with a pair of images in which the only discriminant feature is the location of a dot, an additional dot is added, yielding the EF of proximity or orientation. The EF of linearity is obtained by adding a dot to the orientation pair. These three EFs have been found to elicit strong and consistent grouping effects in humans \citep{PomerantzPortillo}. Each network was tested on 500 randomly generated sets, and each analysis was repeated on three different stroke-over-background conditions (see text).}
  \label{fig:Exp1Stimuli}
  \end{figure}

In order to study individual Emergent Features (EFs) selectively, \cite{PomerantzPortillo} designed an odd-discrimination task in which dot patterns were used to create a base and a composite pair. They found that Configural Superiority Effects (CSEs) were consistently exhibited for three EFs: proximity, linearity, and orientation, which respectively generated a CSE of 0.38, 0.36 and 0.22 seconds (bottom-right box in Figure \ref{fig:Exp1boxplot}), corresponding to a speed-up of RTs of 11\%, 29\% and 26\% \citep{PomerantzPortillo}.
In this experiment, a ``base pair'' consisted of two images with a single dot placed at different locations, whereas  a ``composite pair'' was generated by adding one or two dots in the same location to both base pair images, in such a way as to elicit different emergent properties: orientation, proximity, or linearity (see Figure \ref{fig:Exp1Stimuli}). We generated 500 sets (a base pair and a composite pair for each of the three emergent features tested here). Each dot in the base configuration was constrained to be located at a distance of at least 20 pixels from one another, and 40 pixels from the border in order to avoid border effects \citep{SemihKayhan2020}.

In addition, we also employed a control condition to assess the sensitivity of the networks to the stimuli used in this study (simple features on a randomly pixellated background), due to the highly different appearance from the dataset used for pertaining (which consisted of natural images). To do that, we computed the similarity between a pair of randomly pixellated canvases, and a pair composed of an randomly pixellated canvas and a randomly pixellated canvas with a single dot (see top part of Figure \ref{fig:Exp1Stimuli}). If the pair with the canvas containing a single dot is more discriminable than the pair without the dot, this would imply that the model is indeed sensitive to these types of images.

CEs in humans were measured through an ``odd'' discrimination task and CSEs were established when the added uninformative dot in the composite pair elicited faster discrimination than in the base pair (where only the location of a single dot changed across the two canvasses). In networks, the discriminability of each pair was calculated by presenting each image individually and comparing the internal activation of the networks between the two elements of each pair. If the difference between pairs was greater for the composite (e.g. the ``orientation" pair) than the base pair, this would indicate a superiority effect (CSE), since the uninformative feature added to the composite pair increases the discriminability of the two canvases. Conversely, if the difference between pairs was lower, this would indicate a configural inferiority effect (CIE), where the uninformative feature added to the composite pair \textit{decreases} the discriminability of the two canvases. Unless otherwise stated, the discriminability values are computed through the Euclidean-based metric detailed in Section \ref{MethodMetrics}.

Each experiment was repeated across 3 stroke-over-background conditions: white-over-black, black-over-white, and black-over-randomly pixellated background. We present the results for the latter, but we obtained very similar results across all three conditions.



\subsection{Results}

\newcommand{\mymk}[2]{%
  \tikz[baseline=(char.base)]\node[anchor=south west, line width=0.0pt, draw=none,rectangle, rounded corners, fill=#1, inner sep=2pt, minimum size=0mm, text height=3mm](char){#2} ;}

The results are shown in Figure \ref{fig:Exp1boxplot} for the output layer and in Figure \ref{fig:Exp1_all_layers} for all other layers.  At the output layer, all convolutional networks trained with supervision are sensitive to proximity, linearity, and half of them are also sensitive to orientation. These networks also showed a pattern of responses that mimicked that of human participants (see bottom-right box in Figure \ref{fig:Exp1boxplot}): the proximity feature elicited a bigger effect than linearity, and linearity bigger still than orientation.

\newcommand\tab[1][1cm]{\hspace*{#1}}
\definecolor{lightblue}{RGB}{173,216,230}
\definecolor{wheat}{RGB}{244,222,179}
\begin{figure}[p]
\centering
  \includegraphics[width=1\linewidth]{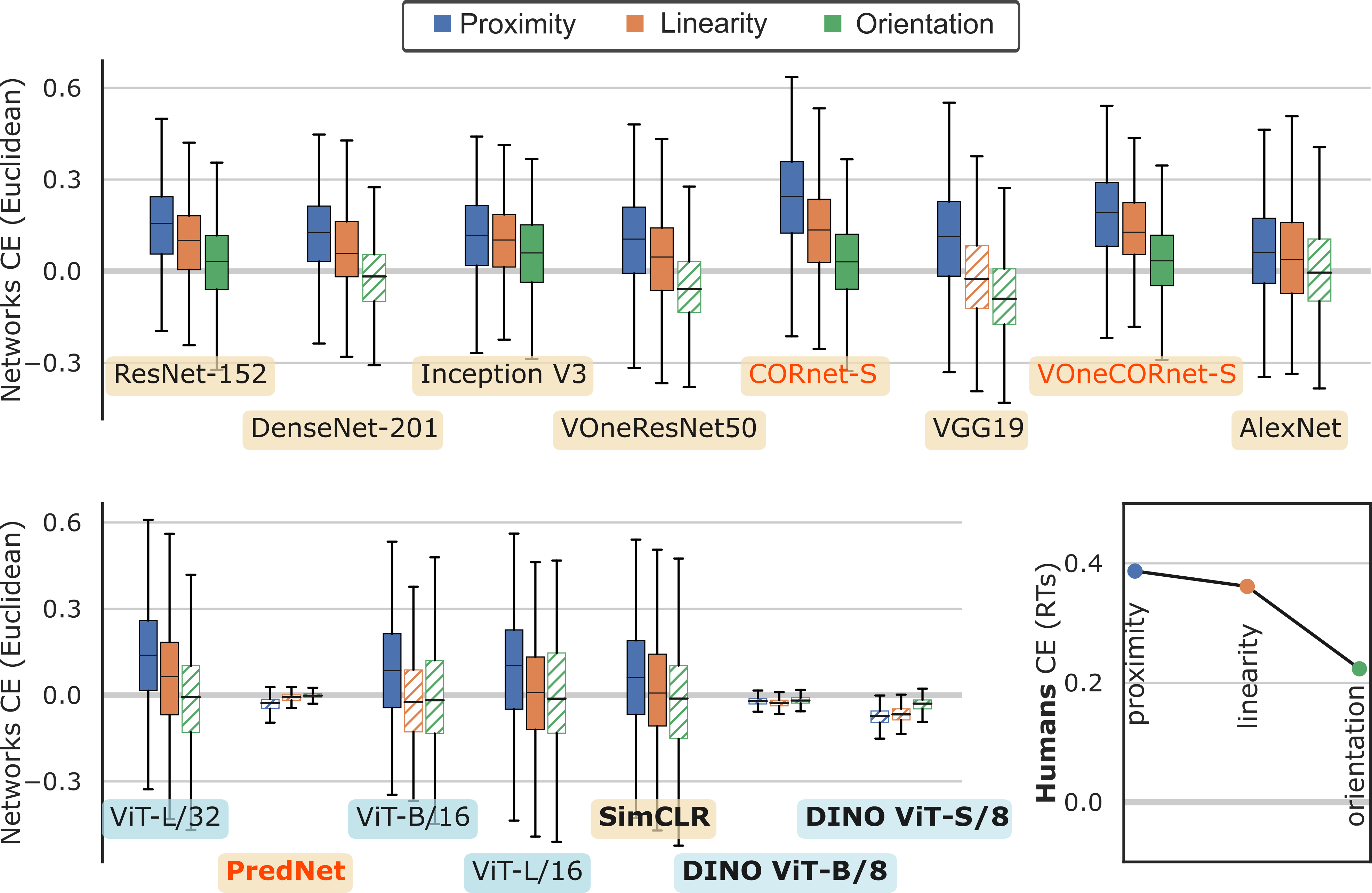}
\caption{Amount of Configural Effect (CE) in Networks and humans (human data showed in the bottom-right box, extracted from \citealt{PomerantzPortillo}). 
In humans, proximity, linearity and orientation all produced high CSEs. Notice that each CE is computed with respect to the base pair and, for the networks, using the Euclidean-based method. 
Most networks not only show consistent CSE across the three EFs, but also a pattern of responses that match the one found in humans. Vision-Transformers showed a less human-like responses, and three out of four self-supervised networks presented a \textit{inferiority} effect. 
All figures refer to the condition with black dots on randomly pixellated backgrounds. 
\newline\tab In this and the following figures, the networks within a  \protect\mymk{wheat}{wheat} colored rectangle are convolutional, those within a \protect\mymk{lightblue}{light blue} rectangle are Vision-Transformers. A normal font-weight indicates supervised learning, a \textbf{bold} font indicates self-supervised, a \textcolor{red}{red} font color indicates the presence of recurrent mechanisms. For example \protect\mymk{wheat}{\textcolor{red}{\textbf{PredNet}}} is a \protect\mymk{wheat}{convolutional}, \textbf{self-supervised} model with \textcolor{red}{recurrence}. In this and the following boxplot figures, each box extends from the first to the third quantile, with a horizontal line at the median. Filled boxes indicate a median $>$ 0, hatched boxes a median $<$ 0. The bars extend from the box by 1.5x the inter-quartile range. Networks are ordered according to their Brain-Score, from the highest to the lowest.}
  \label{fig:Exp1boxplot}
  \end{figure}

\begin{figure}[!ht]
\centering
  \includegraphics[width=1\linewidth]{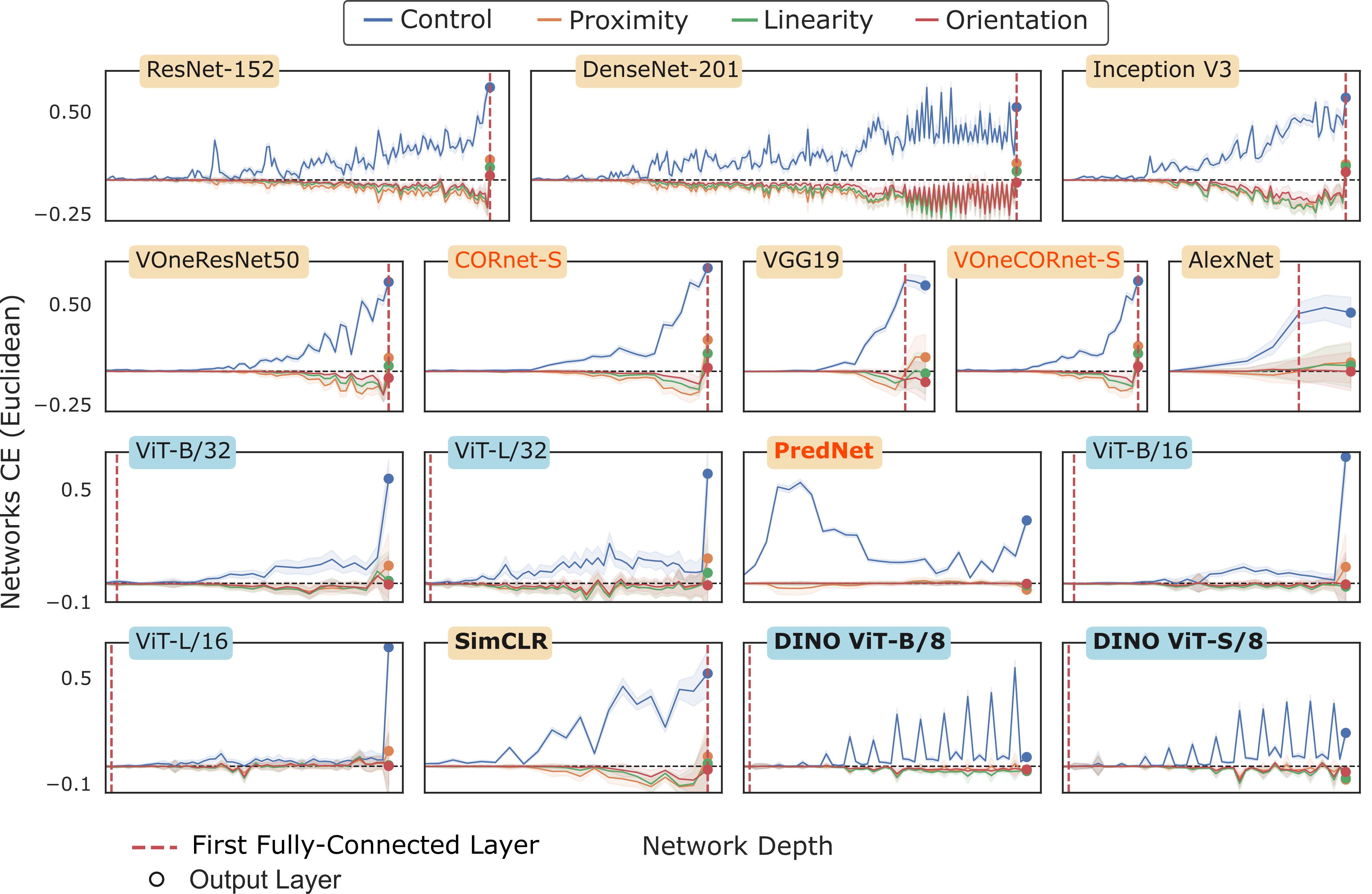}
\caption{Amount of Configural Effect (CE) across all layers, for each network. The control condition (blue) is the only one consistently producing superiority effects across all layers, indicating that the networks are sensitive to the dot stimuli used here. However, the EFs of proximity, orientation, and linearity, produced no effect at early stages and \textit{inferiority} effect at all but the final stages of processing. Dashed vertical red lines indicate the first fully connected layer in each network. Shaded areas correspond to one standard deviation. Networks labels are color-coded as illustrated in the caption of Figure \ref{fig:Exp1boxplot}}
  \label{fig:Exp1_all_layers}
  \end{figure}

Vision Transformers were also always sensitive to proximity, but are inconsistently responsive to the other two features. On the other hand, PredNet and the DINO models showed insensitivity to the three EFs and indeed a \textit{negative} effect (CIE). The very low responses obtained with these networks could be the result of a general insensitivity to the types of dot-stimuli presented in this work. However, the analysis of the control conditions (blue line in Figure \ref{fig:Exp1_all_layers}) shows that all networks used are strongly sensitive to the addition of a single dot starting from the middle stage of processing onwards (with the exception of some Vision Transformers across the middle layers, which showed a weak but still positive effect for the control condition). In all cases, the added dot \textit{increased} discrimination across all networks.
On the other hand, adding one or two dots in order to elicit the EFs of proximity, linearity, and orientation, resulted in a negligible effect for early layers, and a strong but \textit{negative} effect on middle layers, before producing the observed CE at the late stage of processing (see Figure \ref{fig:Exp1_all_layers}). That is, the additional EFs that make the stimuli more discriminable for humans produced \textit{less discriminable} representations in the middle stages of the network. We consider the implication of these results in the General Discussion.  Note, when we repeated the analysis with the cosine similarity approach (as opposed to the Euclidean approach), we found much weaker effects across all networks, and indeed, sometimes the CSEs found in the output layer were reversed into CIEs (see Figure \ref{fig:Exp1boxplot_cossim} in the Appendix). This shows that the CSEs are at least partially encoded in the magnitude of the internal representations.


The results presented thus far refer to the condition with black dots over a randomly pixellated backgrounds. Across the other two background conditions, the results are generally consistent, with the exception of Vision Transformer which showed a higher sensitivity to background, producing more often CIEs for both white and black backgrounds, sometimes even for the proximity condition.


\section{Experiment 2} \label{Exp2Results}
\subsection{Methods}
In Experiment 2 we investigate whether these models show human-like Gestalt phenomena when presented with more complex stimuli taken from \citep{Pomerantz1977PerceptionEffects}.
We used the image pairs from the first two experiments in \cite{Pomerantz1977PerceptionEffects}. Images were arranged in 17 sets, each composed of two pairs: a base and a composite pair (the full set of images, including the added context to each pair, is shown in Figure \ref{fig:Exp2corr}, left). The sets were composed so that they could elicit a wide variety of CSEs and CIEs.
 Furthermore, we used 4 different transformation conditions: no transformation, translation (up to 18\% of the image size), scale (0.7 to 1.3 times the original image size), and rotation (up to 360 degrees). The same random transformation was kept fixed within a set (that is, with a random translation of 20 pixels, both images of the base pair and both images of the composite pair were translated by 20 pixels). For the three conditions employing transformations, each comparison was repeated 500 times, each time with a different random transformation uniformly sampled within the boundaries defined above.

\subsection{Results}
\begin{figure}[!h]
\centering
  \includegraphics[width=0.95\linewidth]{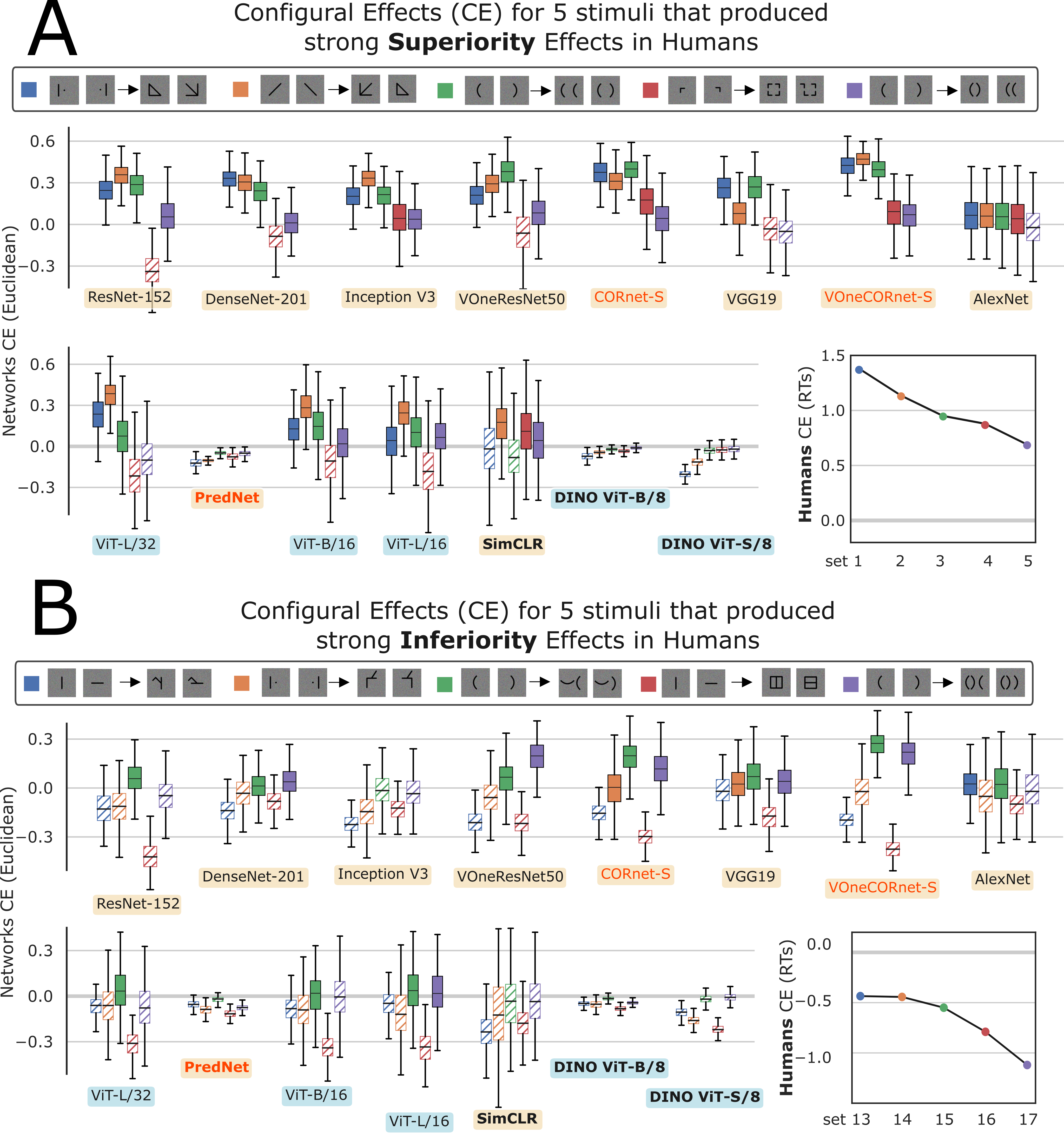}
\caption{\textbf{A:} Networks Configural Effects (CEs) for the output layer for the 5 sets producing high CSEs in humans. All supervised, convolutional networks exhibited CSE for sets 1-3, but overall only three models showed CSE for all stimuli. The pattern did not appear to resemble human responses (in the boxes, data extracted from \cite{Pomerantz1977PerceptionEffects}. Three of the self-supervised networks presented a negative effect. \textbf{B}: Networks' CE for the last layer of the 5 sets producing high inferiority effects (CIE) in humans. Again, the pattern of responses did not match humans', and almost all networks exhibited a superiority effect for at least one of these stimuli. The figure refers to the condition with the randomly pixellated background and translation transformation across pairs. See Figure \ref{fig:Exp1boxplot} caption for details about the boxplots and the color-code used for labelling the networks.}
  \label{fig:Exp2_boxplot}
  \end{figure}

\begin{figure}[!hb]
\centering
  \includegraphics[width=1\linewidth]{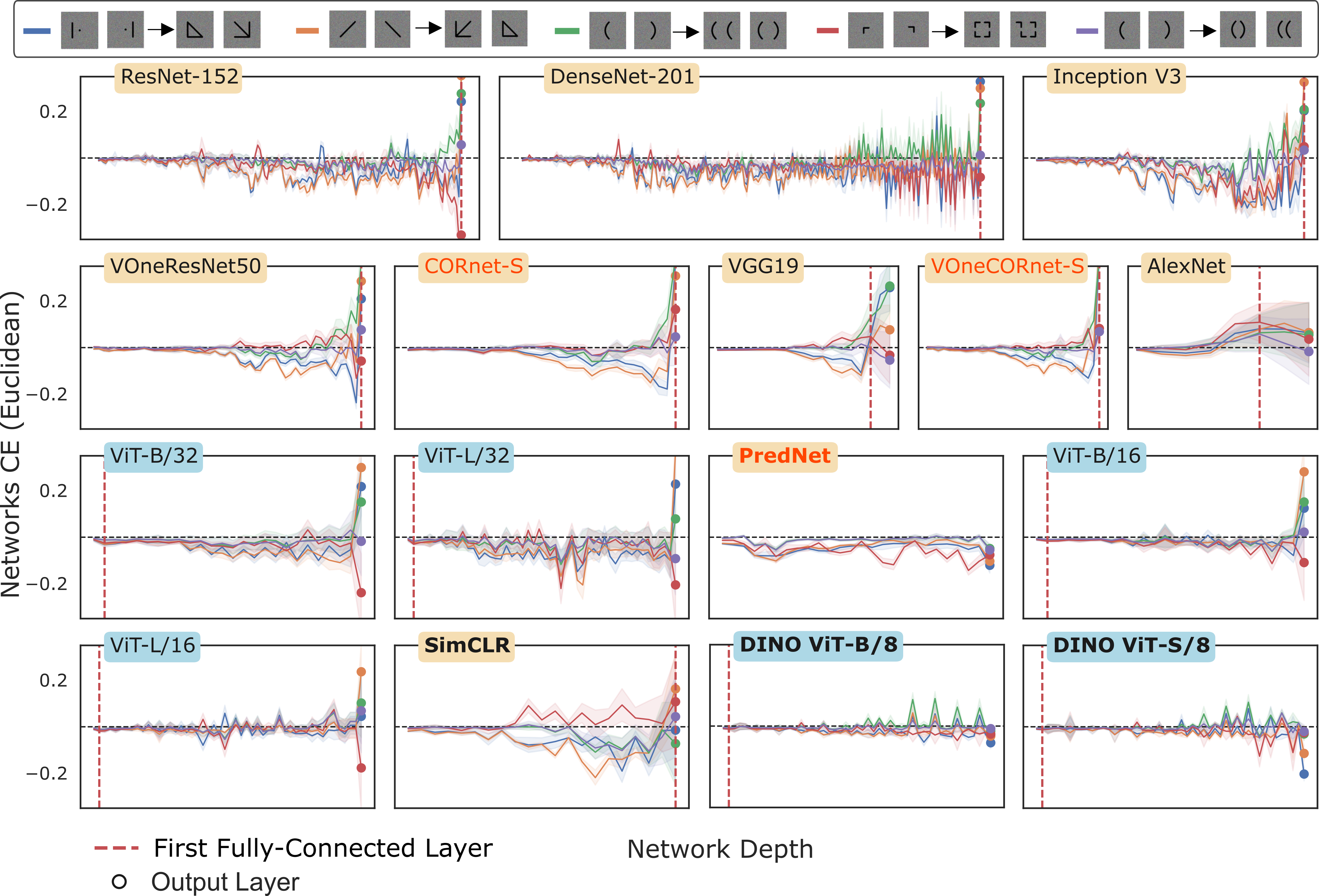}
\caption{Amount of Configural Effect (CE) across all layers, for each network. Similar to the plot in Experiment 1, most networks exhibited a limited sensitivity to Gestalt grouping at early layers, and a mostly negative sensitivity during middle stages, until the output layer. This is in contrast to what expected in human participant (see \nameref{Discussion}). A notable exception was AlexNet and, partially, VGG19. This could be connected with the fact that these networks had fully-connected layers at their middle-late processing stages, and not just at the output layers like all other convolutional networks. Networks labels are color-coded as explained in the caption of Figure \ref{fig:Exp1boxplot}}
  \label{fig:Exp2_all_layers}
  \end{figure}
 Of the 17 sets of stimuli used by \citep{Pomerantz1977PerceptionEffects} we first focus on the 5 sets that showed the largest CSE in humans, corresponding to a speed up between 40\% to 180\% (Figure \ref{fig:Exp2_boxplot}A, the sets are showed in the legend). The corresponding human CSEs are shown in the bottom-right box in Figure \ref{fig:Exp2_boxplot}A. When analysing the output layer, sets 1, 2 and 3 indeed produced positive CSEs for all supervised networks (both convolutional and Transformers). By contrast, sets 4 and 5 were positive only for InceptionNet, CORnet-S, and VOneCORnet-S. And in contrast with Experiment 1, the relationship amongst the CSEs did not resemble that found in humans (for example, in almost all cases, set 2 elicited a stronger effect than set 1). Similarly to the result in Experiment 1, three out of four self-supevised networks presented consistent \textit{negative} effect for all sets (the exception being SimCLR in both Experiments). 
 We then analysed the 5 sets that showed the largest CIE in humans (Figure \ref{fig:Exp2_boxplot}B). We found that not only the pattern of responses did not resemble humans RTs, but almost all networks showed a \textit{superiority} effect for at least one of these sets. 



The analysis across all networks layers (Figure \ref{fig:Exp2_all_layers}) confirmed the pattern observed in Experiment 1, with negligible effects in early layers, and a negative effect in middle layers.

We observed that the two convolutional networks which possessed fully-connected layers in addition to the output (AlexNet and VGG19), demonstrated a CSE across all fully-connected layers, suggesting the possibility that only this type of layer, but not  a convolutional one, may exhibit CSE. We will further explore this point and its connection to studies exploring Gestalt effect using human neural data in the Discussion section.


We extended the analysis to all 17 sets of stimuli by plotting  humans CEs vs the networks CEs in Figure \ref{fig:Exp2corr}, in the Appendix. To quantify whether humans and networks CEs are correlated, we computed Kendall's tau correlation coefficient across human and networks' CEs. The relation was always positive apart for PredNet and DINO ViT-S/8 models, but in no case it was significant at $p<0.01$.

 The results presented thus far refer to the condition with randomly pixelated background and translation transformation. We obtained similar results with the other two background conditions (white and black) and transformation conditions (scale, rotation, and no transformation). When repeating the analysis with the cosine similarity metric we again found an overall reduction of the effect across all layers, which resulted in a larger amount of CIEs.

 \subsubsection{The Effect of Familiarity}
 We investigated whether the CSEs consistently found across some sets of items were the result of familiarity rather than emergent features. That is, it is possible that networks acquired a sensitivity to some basic shapes as a consequence of being trained on ImageNet, and this makes these shapes more salient and discriminable when compared to different shapes or non-shapes. For example, in Set 1 and 2, adding a non-informative stimulus turned the diagonal lines into a familiar geometrical shape (a triangle). In humans, the increase discriminability that results in strong CSEs is assumed to be the result of several emergent features (symmetry, closedness, \citealt{Pomerantz1977PerceptionEffects}). On the other hand, it is possible that networks are sensitive to these images because they have experienced these shapes during training.

We provided an initial test of this hypothesis by modifying sets 1 and 2 (the sets which composite pair included a triangular shape) so that they did not contain familiar shapes anymore. We re-run the analysis only for
those networks that exhibited CSEs for sets 1 and 2. The results (Figure 7) indicated that most networks still exhibited the same degree of Gestalt grouping effects for set 1 and 2 (with the only exception being SimCLR, which
seemed to be sensitive to the particular triangular shape presented in set 2). Therefore, we can reject the hypothesis that the networks were simply sensitive to the familiar shape, and might be sensitive to other features such as closedness.

Overall, Experiment 2 produced mixed results. Only a subset of the stimulus sets that produced large CSEs in humans evoked high CSEs in networks, and almost uniquely at the final layer. In no case the pattern of CSE across stimuli resembled the one produced in humans. Furthermore, networks sometime elicited CSEs in cases where humans most strongly produced CIEs. On the other hand, we verified that networks were not simply basing their responses on familiarity.
 \begin{figure}[!hb]
\centering
  \includegraphics[width=1\linewidth]{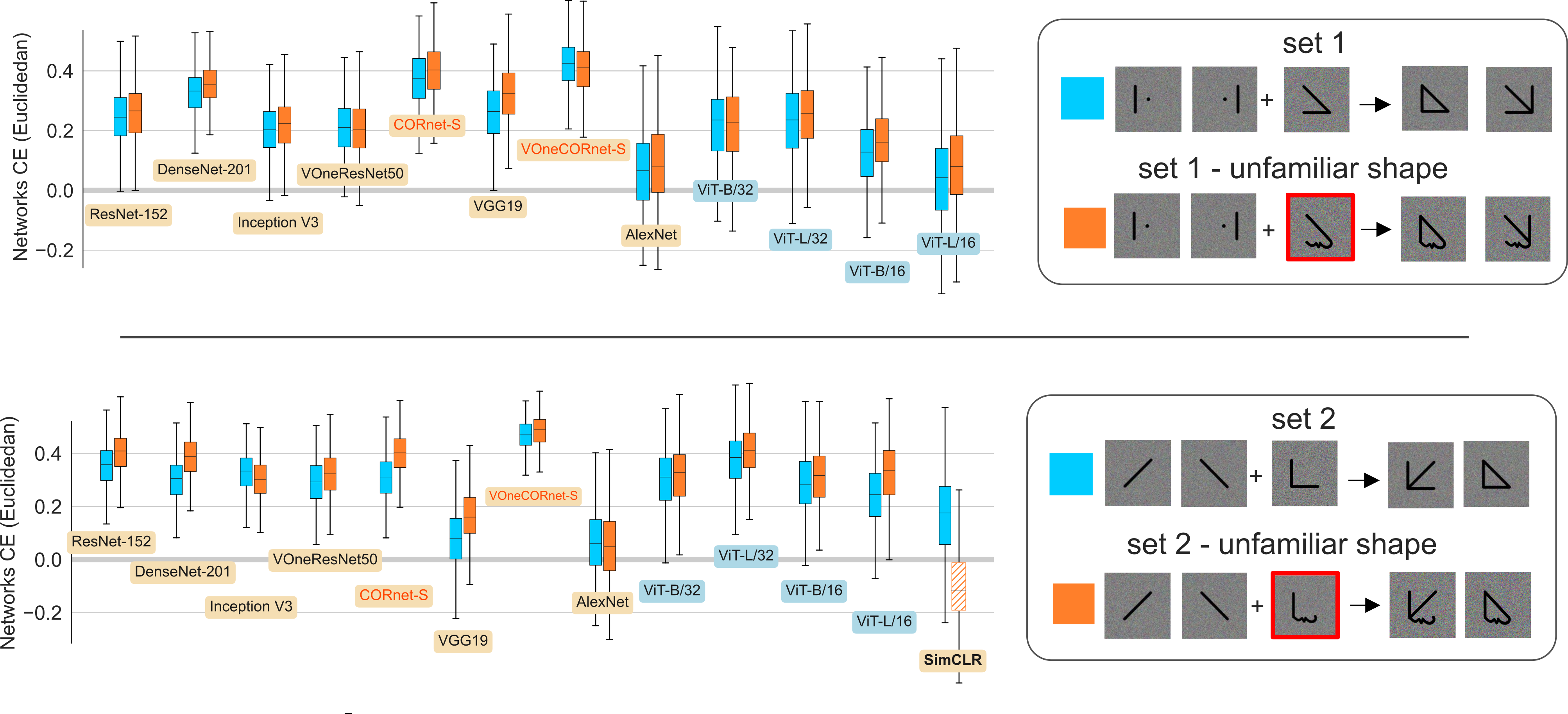}
\caption{Testing the contribution of shape familiarity on sets 1 and 2 by changing the contextual image in order to generate unfamiliar composite pairs. Only networks exhibiting CSEs for these sets were tested. The modified contextual image is shown with a red border. Networks did not seem to be affected by using an unfamiliar shape (with the only exception of SimCLR for Set 2). Therefore, the CSEs found with these shapes is the result of other features, such as closedness. See \ref{fig:Exp1boxplot} for details about the boxplots and the color-code used for labelling the networks.} 
  \label{fig:Exp2shapefam}
  \end{figure}

\section {Discussion} \label{Discussion}
We will now elaborate to what extent our experiments can answer the questions raised in the Introduction and then discuss how our findings relate to previous work assessing Gestalt organizational principles in DNNs.

\subsection{Do modern DNNs exhibit human-like Gestalt grouping?} \label{DiscuDNNgestalt}
Overall, we found mixed support for the hypothesis that DNNs support Gestalt grouping in a human-like fashion. It is indeed remarkable that most convolutional networks trained with supervision did show CSEs for proximity, linearity, and (less frequently) orientation (Experiment 1), and furthermore, the size of the effects mirrored human performance.  However, \textit{for all 16 networks}, the sensitivity to these properties at all layers earlier than the output layer was either negligible or \textit{reversed}.  Furthermore, networks did not show a consistent pattern of responses for complex stimuli: in Experiment 2, only three of the five sets of stimuli that produce CSEs in humans gave rise to consistent CSEs in convolutional networks, and again only at the output layer; the other two showed either a weak effect or the opposite, inferiority effect. Furthermore, in this experiment, the pattern of CSE across different sets did not resemble those observed in human responses. 


EFs are powerful Gestalt properties that are not only subjectively compelling (see Figure \ref{fig:1Pomsetup}) but that support fast RTs in participants, similar to other low-level visual features that ``pop out" \citep{Treisman1998}. The essential role of EFs is reflected in their importance in figure-ground segregation \citep{aCenturyI} as well as building representations of object parts through the perceptual generation of non-accidental properties \citep{Biederman1987, linesKubi2017}. Therefore, it would be expected that EFs emerge in the human visual stream earlier than the stage at which objects are classified. In fact, research has suggested that Gestalt phenomena do not emerge at low-level retinotopic areas, but rather in shape-selective regions such as the lateral occipital complex (LOC) \cite{Kubilius2011}, even though their effect has neural correlates as early as V1 \citep{Fox2017}. Note that LOC corresponds to an earlier processing stage than object identification in the inferotemporal cortex (IT).
This is in striking contrast with what found in this experiment, in which the output layer, often associated with IT, was in most cases the only layer exhibiting any Gestalt effect. Note, we do expect to find CSE in the output layer - but the lack of CSEs (and instead, the strong CIEs) in all earlier layers points to a clear discrepancy between network and human visual processing.

It is possible that the crucial property of the output layer that allows the emergence of CSEs is that it is a fully-connected (FC) layer (as opposed to a convolutional layer). In the current work, amongst the convolutional networks, only AlexNet and VGG19 had multiple FC layers. Interestingly, AlexNet did seem to produce CSE even for ``internal" FC layers in both Experiments, and VGG19 did so for set 1-3 in Experiment 2, but generally failed to produce CSE at all in Experiment 1 even at the output layer (in fact, it was the worst performing convolutional network in that experiment).  

This hints at the possibility that Gestalt effects in DNNs can emerge prior to the output layer and that the reason why effects were restricted to the output layer in most models is that there was only one FC layer. Why fully connected layers but not convolutional layers support some CSEs is not currently clear.  But it suggests that the architectures of most current DNNs are incompatible with the Gestalt organizational principles observed in humans. Furthermore, we noticed that even though the layer taken into account during the analysis of the Vision Transformers were all FC, they did not produce Gestalt grouping consistently, which might be due to the fundamentally different mechanism used by these networks.

\subsubsection{Difference amongst Networks} 
When considering only the output layer, we found a clear difference amongst convolutional and supervised networks (which appeared to be the most sensitive to some basic properties used by humans such as proximity, linearity, and orientation) and self-supervised networks (which produced often the opposite effect as humans). Vision Transformers appeared to be somewhat in between, producing weak and inconsistent human-like effects. It did not appear that networks possessing a recurrent mechanism performed  differently that the other networks. 
Amongst the self-supervised networks, it is surprising that PredNet (a self-supervised, convolutional, and recurrent architecture), which has been shown to be sensitive to several types of illusions \citep{Watanabe2018, Lotter2020}, performed extremely poorly and in fact it always showed inferiority effects. Interestingly, this model also scored poorly on Brain-Score (see Table in \ref{table:networksScore}). On the other hand SimCLR, the lowest scoring model among those used here in the Brain-Score benchmark (almost 3 times lower than the highest scoring model, ResNet-152) did show sensitivity to proximity and linearity, and some CSE for sets 2, 4, and 5 in Experiment 2. Overall, when considering the output layer, we see a clear distinction between networks that were trained with supervision and possessed a convolutional architecture, which showed a higher human-like grouping effect, and all other models. 
However, when considering intermediate layers of processing, the picture was much more homogeneous: most networks' did not show any discrimination between composite and base pairs during early processing, and the two pairs become \textit{even less} discriminable at the middle stages of processing (resulting in CIEs). Almost always networks presented an inferiority effect regardless of architecture, training regime, and their Brain-Score. 

\subsubsection{Magnitude encoding information} \label{DiscussCosineSim}
We observed a small but important difference between the results computed with normalized euclidean distance and cosine similarity (a measure independent of the magnitude of the internal activation). The effects when measured through the latter were strongly reduced and sometime reversed (that is, a response appearing as CSE with Euclidean distance became a CIE with cosine similarity). We believe the Euclidean distance to be a better metric for two reasons: the magnitude of the internal activations is used in the internal computations in the networks, through non-linearities such as ReLU. Furthermore, even if the classification itself is commonly computed by disregarding the last layer' activation magnitude (through argmax), it is nevertheless important to establish that the information is present and could potentially be used if additional layers were to be appended.

\subsection{Acquisition of principles for Gestalt grouping principles} \label{DiscuLearnGestalt}
The degree to which some basic grouping properties can be learnt has been a controversial topic for many decades (see Introduction). Whether or not DNNs are good models of the human visual system, they are excellent at extracting statistical regularities from a dataset, and therefore provide a test to whether some grouping phenomena are implicitly encoded within the statistics of a particular dataset.


The finding that the networks only showed the right pattern of CSEs for low-level features (Experiment 1) suggests that the simplified training environment provided by 2D images might be insufficient for acquiring more complex Gestalt principles, or for combining the low-level features in a human-like way. 
Still, it is possible that using a more realistic dataset (e.g. a 3D environment), or more naturalistic training environments (e.g. with a reward signal such as in the reinforcement learning approach) could result in the acquisition of a wide array of grouping principles, but this is pure speculation for now. On the other hand, we also note that PredNet was trained on a sequence of realistic images (as opposed as static images used for all other networks), and it was  one of the most dissimilar model in terms of showing any type of Gestalt grouping. 



It is also interesting to notice that Gestalt grouping has been obtained in artificial networks that have very different architectures than DNNs \citep{Grossberg1997, Francis2017NeuralCrowding, Herzog2003LocalMasking}. In other cases, architectural modifications have been added to DNNs to solve tasks that appear to underlie Gestalt grouping \citep{Linsley2018}. Highly specific and unnatural training setups (e.g. training on a checkerboard-like pattern) also resulted in the emergence of some configural properties in convolutional DNN \citep{Configural2021}. This suggests that  Gestalt principles might be obtained only with appropriate architecture or with highly specific training, and are not simply the result of statistical feature extraction from the retinal images. 


\section{Conclusion}
Overall, the results presented here highlight a partial disconnect between DNNs trained on natural 2D images and human vision. Testing 16 networks that varied across architectures and training setups on stimuli that elicited strong Gestalt grouping (measure as increased discrimination in an ``oddity Reaction Time task") in humans, we found that all models almost universally showed no grouping effect across early layers, and a \textit{negative} effects (meaning a decreased discrimination) across middle and late stages of processing up to the output layer. This is in contrast to the human visual processing system, where the effect of emergent features has been observed at intermediate stages of processing. 
At the output layer, only  convolutional method trained with supervision could consistently acquire low-level visual features that elicited grouping effects that strongly resembled those produced by humans.   
At the same time, the sensitivity to low-level grouping features in trained networks did not fully to transfer to more complex stimuli, which instead produced grouping effects that were only inconsistently mimicking human behaviour and were sometime strongly deviating from it. Although our findings highlight the limitations of current DNNs in capturing Gestalt effects, our results do suggest the possibility that human-like Gestalt effects may emerge in response to training in some network architectures. More generally, this work highlights the importance of comparing network performance with well-established psychological phenomena, which have been largely ignored when comparing DNNs to the human brain.


\section {General Methods}  \label{Methods}


\subsection{Network Used} \label{method networks}
We selected 16 networks based on their historical importance, their performance on standard datasets, and their biological plausibility. We used as a point of reference the Brain-Score value \citep{SchrimpfBrainScore2018}, indicating the amount of variance explained by the model across several benchmarks.
\textbf{AlexNet} \citep{NIPS2012_c399862d}, \textbf{VGG19} \citep{VGG16}, and \textbf{ResNet} \citep{He2016} (we used ResNet-152) are classic networks that have been often tested on several cognitive phenomena \citep{SchrimpfBrainScore2018, Baker2018CogSci, Biscione2022NN}, with mixed results.  
\textbf{InceptionNet} \citep{InceptionNet2015} was shown by \cite{Kim2019} to be sensitive to the effect of Gestalt Closure and thus it seemed suited for this battery of tests (we used InceptionNet V3). 
In \textbf{DenseNet} \citep{Huang_2017_CVPR} each convolutional layer is connected with each other layer. A smaller version of this family of networks, which we used (Densenet-201), has been shown to possess human-like translation invariance \citep{Biscione2021JMLR}.


We also tested two models specifically developed to be biologically plausible and that provide a good match with primate neural data. 
The ``CORnet" model family \citep{CorNet} aimed to incrementally build a network architecture by adding recurrent and skip connections while monitoring both classification accuracy and agreement with a body of primate brain neural data. From this family, the \textbf{CORnet-S} was selected as the best CORnet architecture. 

The VOneNet family \citep{Dapello2020} has been developed to better match the structure of the primate visual cortex. Each VOneNet contains a fixed-weight neural network front-end that simulates primate V1, called the VOneBlock, followed by a neural network back-end adapted from current CNN vision models. We used two versions of VOneNet: one with CorNet-S backend (\textbf{VOneCORnet-S}), and the other with Resnet50 backend (\textbf{VOneResnet50}). 

Vision Transformers \cite{Mehrer2021AnVision} are an attention-based family of models which achieve higher accuracy on vision tasks and have been found to also have a more consistent pattern of error with those of humans \cite{Tuli2021}. We used \textbf{ViT-B/16}, \textbf{ViT-B/32}, \textbf{ViT-L/16}, \textbf{ViT-L/32}, where B/L indicates either a ``base`` or a ``large`` model, and the number indicates the patch size. In spite of their success in vision tasks, Vision Transformer do not have a particularly high success with the Brain Score 

Self-supervised models aim at building meaningful representations from unlabelled data that can be used for solving downstream tasks, and the deep contrastive embedding they use has been suggested to be a biologically-plausible computational theory of the primate visual system \citep{Zhuang2020}. We used four self-supervised networks: \textbf{Dino ViT-S/8}, \textbf{Dino ViT-B/8} \citep{CaronDINO2021}, \textbf{SimCLR-ResNet18} \citep{chen2020simple} and \textbf{PredNet} \citep{Lotter2018}. The name after the space indicates the networks' backbone. PredNet is particularly interesting for this work as it is inspired by the predictive coding theory in the neuroscience literature \citep{Rao1999} and can mimic several effects of visual perception, including illusory contours, the flash-lag effect \citep{Lotter2017} and illusory motion \citep{Watanabe2018}.

VOneNet-Resnet50, DenseNet, ResNet-152 and VGG19 are in the top 10 on Brain-Score at the moment of writing, all with a score higher than $0.4$ (the highest scoring network with $0.465$ was a version of EffNet which we were not able to obtain). Amongst the convolutional, supervised models, AlexNet scored the lowest with $0.38$. In spite of their success in vision tasks, Vision-Transformer do not have a particularly high success with the Brain Score (ViT-B/32 scored the highest with $0.355$, all other ViT scored between $0.16$ and $0.2$). Similarly, SimCLR score is as high as AlexNet, which scored the lowest amongst convolutional supervised networks ($\sim 0.38$).  Surprisingly, PredNet is the lowest scoring models amongst the one tested here ($0.195$). All values are shown in  \ref{table:networksScore}.  

All networks were pretrained on ImageNet. When feeding the images into the networks, we first resized them to the same size used during the networks' pretraining (that is $224$x$224$ for all networks but InceptionNet, which was $299$x$299$). Furthermore, all images were normalized with mean and standard deviation used during ImageNet pretraining. We analysed the activation of all Convolutional and Fully Connected layers before the non-linearity operation was applied.

\subsection{Computing Network Configural Effets} \label{MethodMetrics}

The main difficulty in comparing Pomerantz's behavioural results with neural networks is that the behavioural results were based on RTs, which DNNs do not produce (there are some exceptions, but the models commonly scoring high on Brain-Score do not possess this feature). However, since we have direct access to models' internal representations, we can nevertheless obtain a measure of stimuli discriminability. We presented a pair of images to the network; for each image, we recorded the value of activation for every unit of a given layer, obtaining an activation vector for each image and each layer. The ``distance" between the two activation vectors would correspond to a measure of discriminability for the image pair at a particular layer. We compared these measures to human RTs: high distance would correspond to high discriminability, which would correspond to fast RTs; and vice-versa.

We used two different ways of comparing the distances across the activation vector $d^l(\textbf{x})$ (where $l$ indicates a specific layer, and x is an input image): a Euclidean-based method:
$$ D^l(\textbf{a}, \textbf{b}) = \left \| d^l(\textbf{a}) - d^l(\textbf{b}) \right \|,$$

and a cosine similarity metric, which is invariant to the magnitude of the internal activations: $$C^l(\textbf{a}, \textbf{b}) = \frac{d^l(\textbf{a}) \cdot d^l(\textbf{b})}{\left \| d^l(\textbf{a}) \right \| \left \| d^l(\textbf{b}) \right \|}.$$

 We refer to the general difference across composite and base as Configural Effect (CE). By using the same approach outlined in \cite{PomerantzPortillo}, we obtained the networks' Configural Superiority Effects (CSEs) and Configural Inferiority Effects (CIEs) by computing the difference for each distance metric across the two pairs of stimuli: a base and a composite pair. The composite pair is obtained by adding a non-informative feature to each image of the base pair.
For humans, CE is simply computed as $RT_{base} - RT_{composite}$, with positive values indicating CSE and negative CIE.

For networks, CEs with the Euclidean distance can be computed as follows \cite{Jacob2021}: 
$$
NetworkCE = \frac{D^l(composite_a, composite_b) - D^l(base_a, base_b)}{D^l(composite_a, composite_b) + D^l(base_a, base_b)}
$$
Since $D^l$ is a measure of \textit{dissimilarity}, a $D^l$ higher for the composite pair than for the base pair indicates that the uninformative feature added to the composite pair made the two images more dissimilar (that is, more discriminable) from one another, which would correspond to CSE. 

Alternatively, we can compute the CE using cosine similarity: 
$$NetworkCE = C^l(base_a, base_b) - C^l(composite_a, composite_b).$$
Being the cosine similarity a measure of \textit{similarity}, a base pair more similar than the composite pair will again indicate CSE. In both cases, positive $NetworkCE$ indicates CSE, and negative $NetworkCE$ indicates CIE.

Note that the Euclidean and cosine similarity approaches to computing CE will have different scales, which will again be different than humans CE computed through RTs.

We compute these metric for each layer in the networks, in the same order in which they are traversed during the feedforward pass. That is, for the networks in this work using recurrence (CORnet-S, VOneCORnet-S, PredNet) we show the distance at the recurrent layers multiple times, one for each time it's been traversed, and in the same order it's been computed.

\section*{Declarations}

\paragraph{Code Availability}
Code is provided in full: \url{https://github.com/ValerioB88/gestalt-DNNs}. 

\paragraph{Data Availability} In the code, stimuli are generated at runtime, and therefore no dataset is needed to replicate the results when using our code. However, a dataset of the stimuli used in the experiments is provided: \url{https://valeriobiscione.com/PomerantzDataset}. 

\paragraph{Author Contribution} Both authors contributed ideas for the project. VB created the datasets, ran the experiment and the analysis. Both authors contributed to writing the manuscript.
\paragraph{Funding} This project has received funding from the European Research Council (ERC) under the European Union’s Horizon 2020 research and innovation programme (grant agreement No 741134).

\paragraph{Ethics Approval} Not applicable
\paragraph{Consent for Publication} Not applicable
\paragraph{Consent to Participate} Not applicable
\paragraph{Conflict of Interests} The authors declare no competing interests.

\bibliographystyle{elsarticle-harv} 
  \bibliography{references}


\appendix
\section{Cosine Similarity Analysis} \label{Appendix1}
\begin{figure}[!ht]
\centering
  \includegraphics[width=1\linewidth]{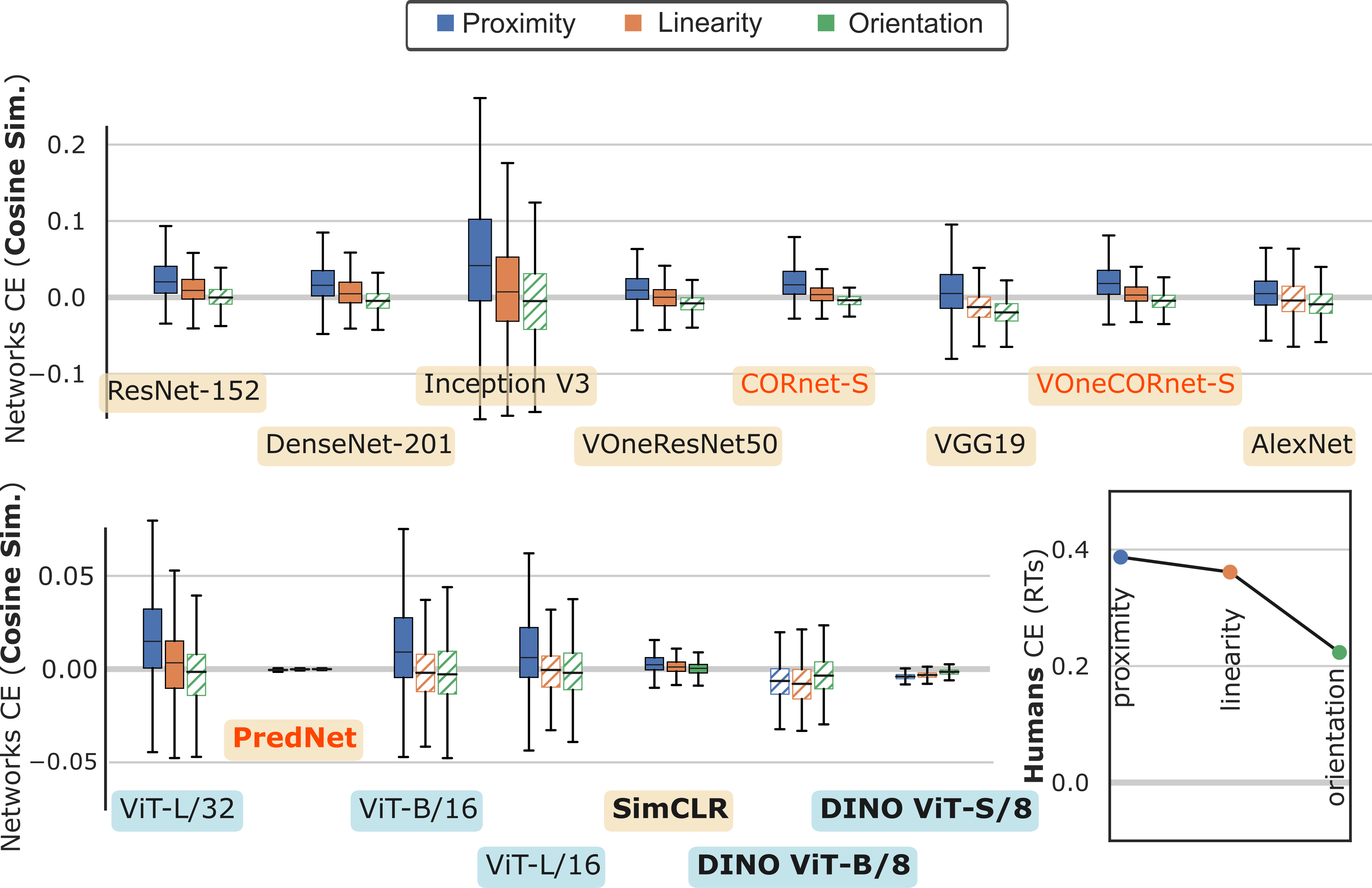}
\caption{Result from Experiment 1 with cosine similarity approach as opposed to the Euclidean-based method presented in the main text. Amount of Configural Effect (CE) in Networks and humans (bottom-right box, data extracted from \citealt{PomerantzPortillo}).  Even though the pattern of responses is similar across humans and networks (that is, the order from the most to the least CE is proximity, linearity, and orientation), DNNs often exhibited CIE, in contrast with the Euclidean-based approach in which they express CSE more frequently, especially the convolutional architectures. No networks but SimCLR showed CSE for orientation, and many networks failed to show CSE for linearity.} 
  \label{fig:Exp1boxplot_cossim}
  \end{figure}

We report here the cosine similarity results for the last fully connected layer for Experiments 1 and 2 in the paper. Figure \ref{fig:Exp1boxplot_cossim} shows the result for Experiment 1. Overall, the cosine similarity appears to reduce the amount of CSE in such a way that some EFs went from CSE to CIE (that is, adding an additional dot now \textit{hindered} recognition). The cosine similarity revealed CSE to proximity for all networks, it always results in CIE for orientation, and showed more mixed results for the linearity condition.
The pattern for Experiment 2 is shown in Figure \ref{fig:Exp1boxplot_cossim}. Like before, the cosine similarity analysis underestimates the amount of CSE present compared to the Euclidean-based metric, and more sets result in CIE compared to the Euclidean-based approach. 
However, the overall picture is the same for both metrics: networks produce CSE in a very inconsistent way, the shape of CSE is dissimilar to those of humans, and it sometimes produces CSE with stimuli that strongly elicit CIE in humans (Figure \ref{fig:Exp1boxplot_cossim}, bottom). For the layers earlier than the output ones, using cosine similarity also decreased the amount of CE across all networks but, as discussed in the main text, these were already in the great majority CIEs.

Since cosine similarity does not take the magnitude of the activation vector into account, and that cosine similarity consistently under-estimate CSEs, we can conclude that the magnitude of the internal activation vector contributes in encoding the effect of the EFs.

\begin{figure}[!ht]
\centering
  \includegraphics[width=1\linewidth]{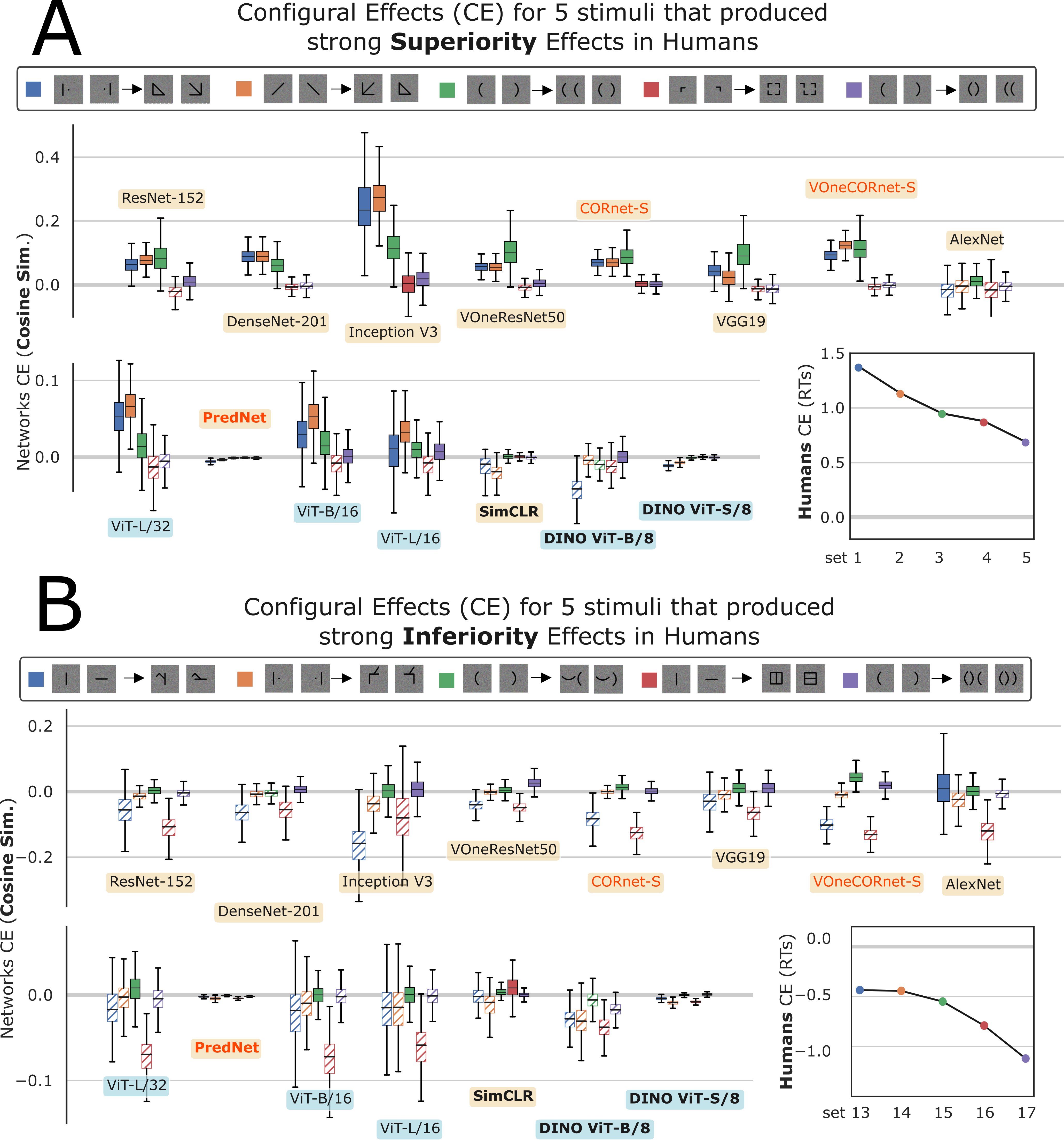}
\caption{Result from Experiment 2 with cosine similarity approach as opposed to Euclidean-based method. Amount of Configural Effect (CE) in Networks and humans (boxes, data extracted from \citealt{PomerantzPortillo}), for sets eliciting strong CSE (\textbf{A}) and strong CIE (\textbf{B}). sets 1-3 showed a consistent effect (similar to what was found with Euclidean distance). As before, there is an overall decrease in CE resulting in CIE in more cases than when using the Euclidean method. Overall the pattern of responses is strongly dissimilar to those found in humans. It is however still noteworthy that networks do show a certain amount of Gestalt grouping for some sets.} 
  \label{fig:Exp2boxplot_cossim}
  \end{figure}

\newpage
\section{Experiment 2: All-Stimuli Correlation}
\begin{figure}[!h]
\centering
  \includegraphics[width=1\linewidth]{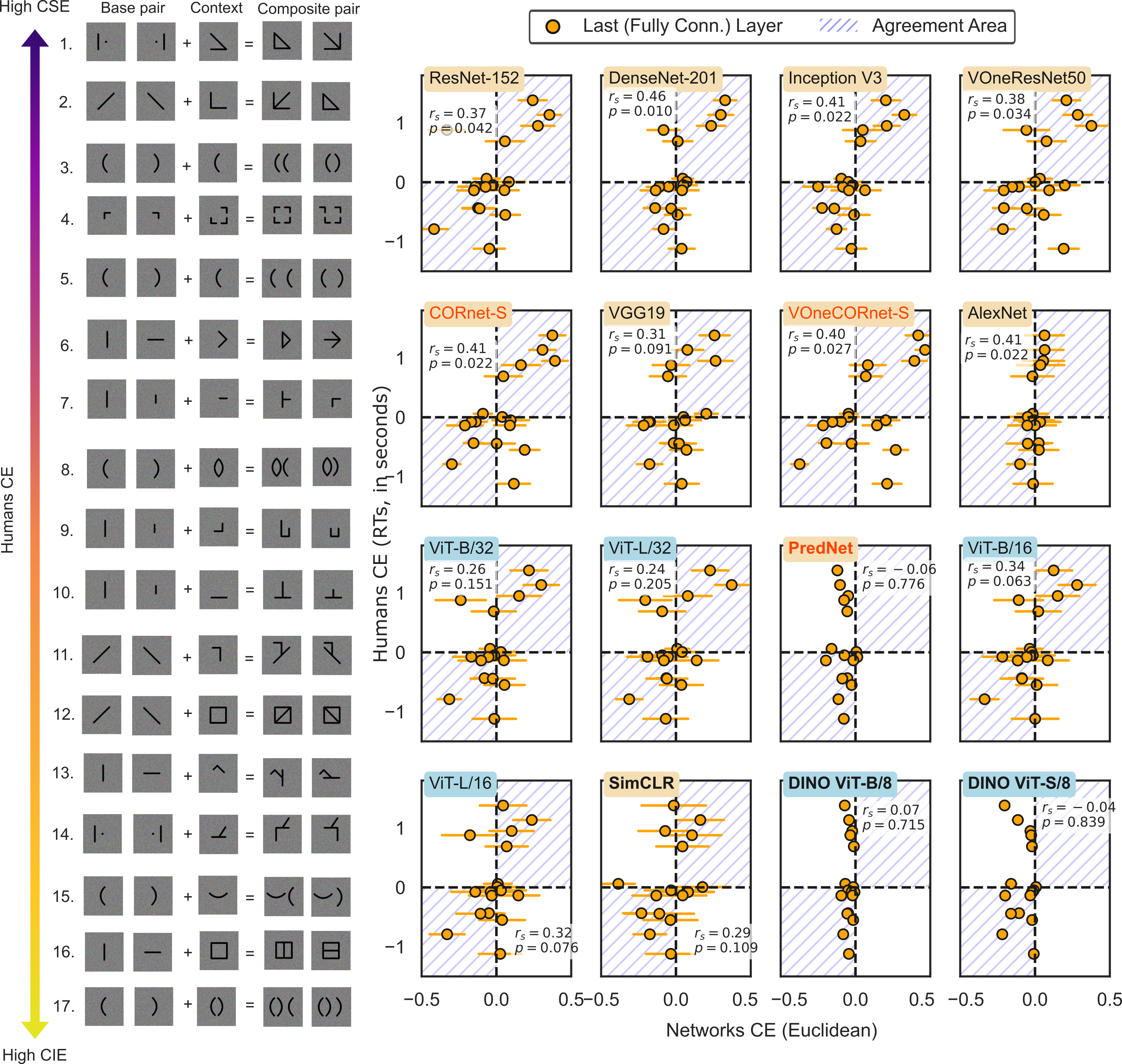}
\caption{\textbf{Left}: the full set of base and composite pairs used in the experiment to assess Configural Effects (CE) in networks. sets are sorted by the amount of CEs elicited in humans, measured in \cite{Pomerantz1977PerceptionEffects}. Gestalt grouping is measured by CSE using the Euclidean-based method. \textbf{Right}: networks and humans' CE for the 17 sets, at the last layer (black-on-random-pixel background, translation condition). For each subplot, points in the top-right area indicate human and network agreements on Gestalt grouping; points in the bottom-left area indicate agreement on crowding/ context interference. The Kendall's rank-correlation coefficient $r_s$ for the last layer indicates a moderate agreement (orange circles).}
  \label{fig:Exp2corr}
  \end{figure}

While in the main text we focused on the five stimuli that produced the most CSE and CIE, here we show the result across all 17 stimuli used in Experiment 2 (Figure \ref{fig:Exp2corr}) at the output layer. The stimuli are recreated according to \cite{Pomerantz1977PerceptionEffects}. The human CE are computed from the same source. Network CE are computed using the Euclidean-based approach. 

\null\newpage
\section{Appendix C. Brain-Score for Network Used} \label{networksScore}
We show here the Brain-Score averaged across all benchmarks for each of the networks used in the present work. When a network has been tested multiple times and has multiple scores, we indicated the highest one. Collected on January 31, 2023.
\begin{table}[h] \label{table:networksScore}
\begin{tabular}{|l|l|}
\hline
\textbf{Network Name} & \textbf{Average Brain Score} \\ \hline
ResNet-152            & 0.432                        \\ \hline
DenseNet-201          & 0.421                        \\ \hline
Inception V3          & 0.414                        \\ \hline
VOneResNet50          & 0.405                        \\ \hline
CORnet-S              & 0.402                        \\ \hline
VGG19                 & 0.402                        \\ \hline
VOneCORnet-S          & 0.390                        \\ \hline
AlexNet               & 0.381                        \\ \hline
ViT-B/32              & 0.355                        \\ \hline
ViT-L/32              & 0.198                        \\ \hline
PredNet               & 0.195                        \\ \hline
ViT-B/16              & 0.190                        \\ \hline
ViT-L/16              & 0.161                        \\ \hline
SimCLR-ResNet18       & 0.160                        \\ \hline
DINO-ViT-S/8          & -                            \\ \hline
DINO-ViT-B/8          & -                            \\ \hline
\end{tabular}
\end{table}

\end{document}